\documentclass[10pt,twocolumn,letterpaper]{article}

\usepackage{authblk}
\usepackage{cvpr}              
\usepackage[accsupp]{axessibility}
\usepackage{graphicx}
\usepackage{amsmath}
\usepackage{amssymb}
\usepackage{booktabs}
\usepackage[dvipsnames]{xcolor}

\usepackage{lipsum}
\usepackage{enumitem}
\usepackage{multirow}

\makeatletter
\renewcommand\AB@affilsepx{\qquad \protect\Affilfont}
\makeatother

\newcommand\Tstrut{\rule{0pt}{2.2ex}}         

\usepackage[pagebackref,breaklinks,colorlinks]{hyperref}

\usepackage[capitalize]{cleveref}
\crefname{section}{Sec.}{Secs.}
\Crefname{section}{Section}{Sections}
\Crefname{table}{Table}{Tables}
\crefname{table}{Tab.}{Tabs.}

\def\cvprPaperID{17} 
\def\confName{CVPR}
\def\confYear{2022}

\begin{document}

\title{\vspace{-30pt}GCA-Net : Utilizing Gated Context Attention for Improving \\ Image Forgery Localization and Detection \vspace{-20pt}}

\author[$1\dag$]{Sowmen Das}
\author[$2\mathparagraph$]{Md. Saiful Islam}
\author[$3\mathsection$]{Md. Ruhul Amin}

\affil[\space]{\quad $^\dag$Shahjalal University of Science and Technology, Bangladesh \protect \\}
\affil[\space]{\hspace{-10pt}$^\mathparagraph$University of Alberta, Canada \quad}
\affil[$\mathsection$]{Fordham University, USA \protect \\}
\affil[ ]{\hspace{-30pt} \textit {\normalsize $^1$sowmendipta@gmail.com \quad $^2$mdsaifu1@ualberta.ca \quad $^3$mamin17@fordham.edu}}

\maketitle

\begin{abstract}
   Forensic analysis of manipulated pixels requires the identification of various hidden and subtle features from images. Conventional image recognition models generally fail at this task because they are biased and more attentive towards the dominant local and spatial features. In this paper, we propose a novel Gated Context Attention Network (GCA-Net) that utilizes non-local attention in conjunction with a gating mechanism in order to capture the finer image discrepancies and better identify forged regions. The proposed framework uses high dimensional embeddings to filter and aggregate the relevant context from coarse feature maps at various stages of the decoding process. This improves the network's understanding of global differences and reduces false-positive localizations. Our evaluation on standard image forensic benchmarks shows that GCA-Net can both compete against and improve over state-of-the-art networks by an average of $4.7\%$ AUC. Additional ablation studies also demonstrate the method's robustness against attributions and resilience to false-positive predictions.
\end{abstract}
\section{Introduction}

\begin{figure}[t]
\centering
  \includegraphics[width=0.98\linewidth]{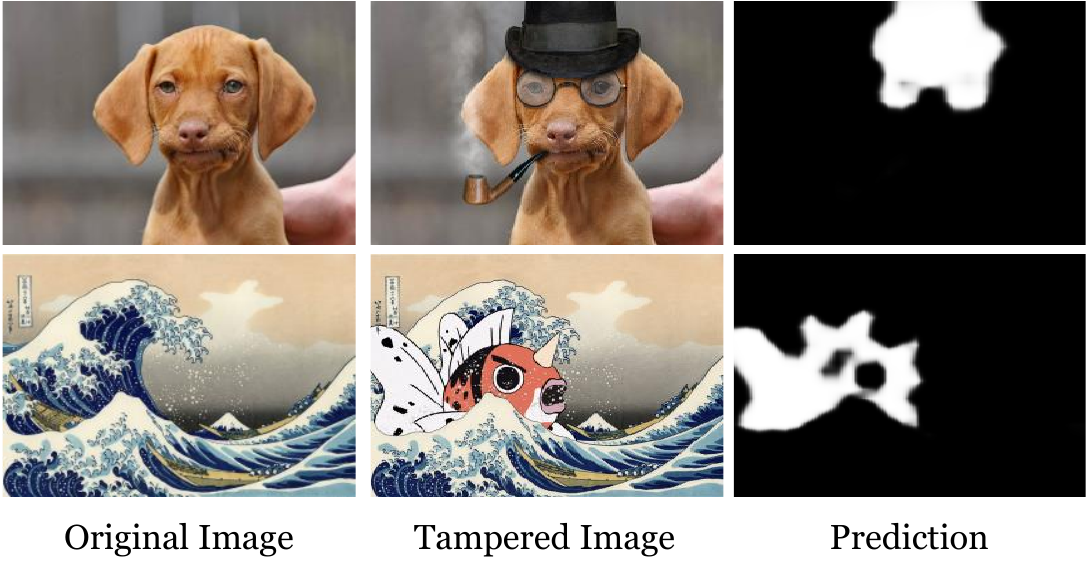}
   \caption{The first and second columns contain examples of authentic and manipulated images. The third column shows the output generated by our proposed network, highlighting the manipulated regions of the tampered images.}
\label{fig:sample}
\end{figure}

Image manipulation is the act of altering an image's content using different editing techniques. Examples of such manipulation include, content removal \cite{inp1}, face-swapping \cite{faceswap}, attribute changing \cite{attributechanging}, \textit{etc}. There are three primary types of manual image forgery techniques \cite{imanip_survey} $-$ splicing, copy-move, and inpainting. The majority of these manipulation scenarios consist of copying and pasting some pixels from a source image onto a target image, as shown in. Fig. \ref{fig:sample}. Forensic algorithms play the important role of determining whether an image is authentic or not. Further research in developing better detection mechanisms is critical in today's age of social media, as any fake news can spread rapidly and be used to foment panic and propaganda.

Numerous methods have been proposed over the years for image forgery localization and detection (IFLD).  In recent years the use of deep CNN architectures capable of learning intrinsic features has gained popularity \cite{detect_survey}. However, unlike traditional image classification, forgery detection involves identifying hidden manipulation traces from the image rather than the apparent spatial content.
Existing works that utilize a variety of sequential frameworks \cite{mantranet, dual_domain, ghosh2019spliceradar} suffer from feature attenuation due to the subtlety of the forgery artifacts. Although some methods try to utilize feature hierarchies through recursive pooling \cite{span, lstm}, they are mostly bound to local neighborhoods. 

\begin{figure}[t]
\centering
  \includegraphics[width=0.98\linewidth]{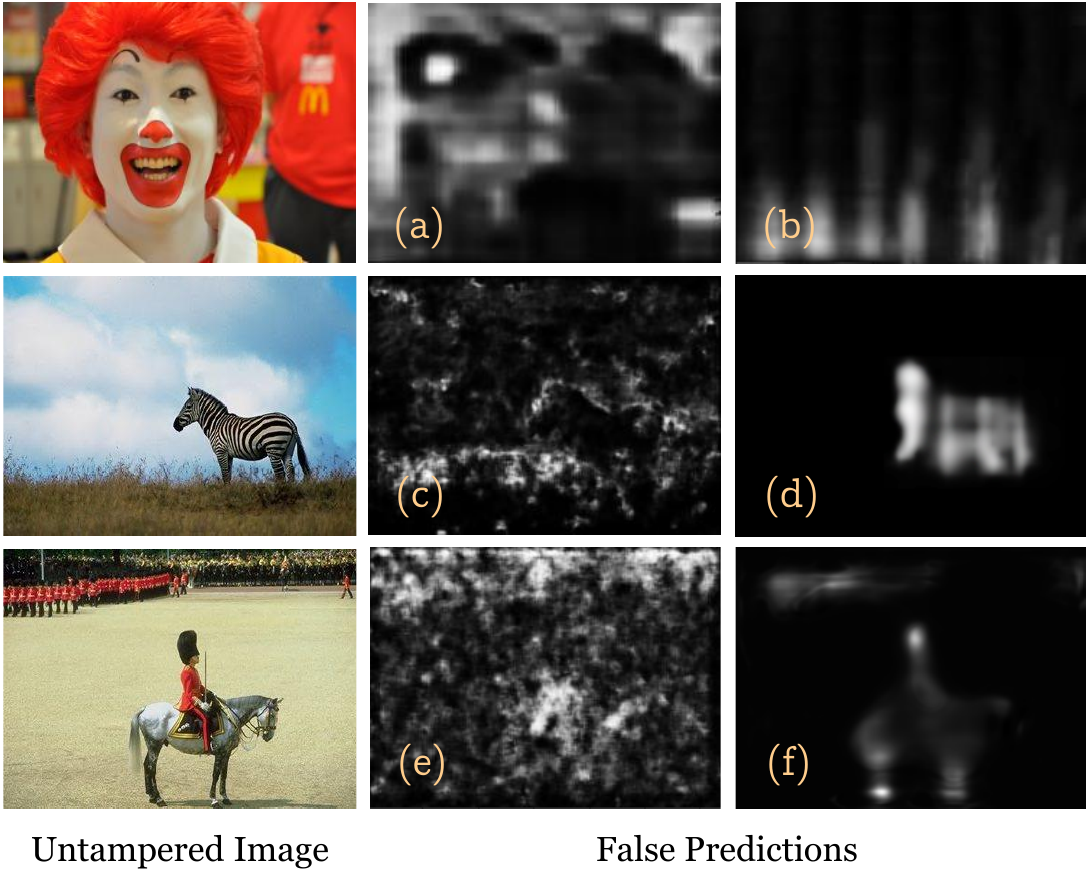}
   \caption{Examples of false predictions from different networks. The first column contains authentic images and the next two contain results of the following networks $-$ (a) Bappy \etal \cite{lstm}, (b) Zhou \etal \cite{zhou2019generate}, (c,e) Wu \etal \cite{mantranet}, (d) Wu \etal \cite{deepmatching}, (f) Wu \etal \cite{busternet}. We can see that the networks highlight a lot of erroneous pixels for untampered images. }
\label{fig:auth}
\end{figure}

Majority of the existing manipulation localization networks \cite{span, mantranet, noiseprint} perform a pixel-level binary classification of each individual pixel in order to generate the localization map. Image-level classification is consequently done depending on the percentage of identified pixels. This simplification results in two notable drawbacks $-$ i) Since these methods view each pixel as a single data point, they fail to utilize important region characteristics such as noisy boundaries and object artifacts that can help in discerning region differences, ii) Despite authentic images not containing manipulation traces, methods that were trained using this process exhibit erroneous predictions when trying to localize untampered images; shown in Fig. \ref{fig:auth}. 


In order to address the shortcomings of existing networks, we propose the Gated Context Attention Network (GCA-Net), which is composed of an improved feature encoder and a densely connected attention decoder. Image manipulation alters not only the visible content of an image but also the underlying signal and noise characteristics. So, in order to utilize these multi-modal features, we combine EfficientNet \cite{tan2020efficientnet} with an improved Error Level Analysis (ELA) \cite{ela} module and other steganalysis layers to identify the compression and noise artifacts. However, since every image contains some intrinsic noise patterns \cite{noiseprint}, in order to localize the manipulations, we must correlate the local noise difference to the global image fingerprint. Although previous works \cite{mantranet, span, rich_features} have attempted to use these signal features, their inattention to the global similarities resulted in generating noisy false positives.

Our proposed GCA-Net uses a novel gated context attention which performs two distinct computations. Firstly, it aggregates multiple cross-layer feature embeddings and applies a global context attention to identify differences between local and global representations. This improves the network's understanding of long-range dependencies between the various pixel regions. Secondly, the low-level context features are filtered using an attention gate to eliminate redundant spatial and object data. By propagating only the necessary forensic features, we are able to minimize attenuation and improve the identification of boundary artifacts. Furthermore, the dense network structure of Unet++ \cite{unet++} facilitates information exchange between intermediate layers, thus reducing feature loss and improving convergence. This design also increases the network's robustness to post-processing attributions. Our evaluation on the popular CASIA \cite{casia}, NIST \cite{nist}, and IMD \cite{imd2020} benchmarks show performances comparable to existing SOTA methods under constrained resources. GCA-Net outperforms several existing methods by $4\% - 6\%$, successively generating better localization and fewer false positives. To summarize the contributions of our work:
\setlist{nolistsep}
\begin{itemize}[leftmargin=*]
    \item We propose GCA-Net consisting of a novel Gated Context Attention module that enables efficient modelling of long-range dependencies and improves global context representation necessary for manipulation localization.
    \vspace{1.2pt}
    \item We substantiate the impact of global context and non-locality in detecting image forgeries, facilitating future research in this field.
    \vspace{1.2pt}
    \item We illustrate the effects of training strategies for reducing false positives and improving image-level predictions.
    \vspace{1.2pt}
    \item We show that our method outperforms the state-of-the-art performance on standard datasets for image manipulation detection and localization.
\end{itemize}

\begin{figure*}
\centering
\includegraphics[width=\linewidth]{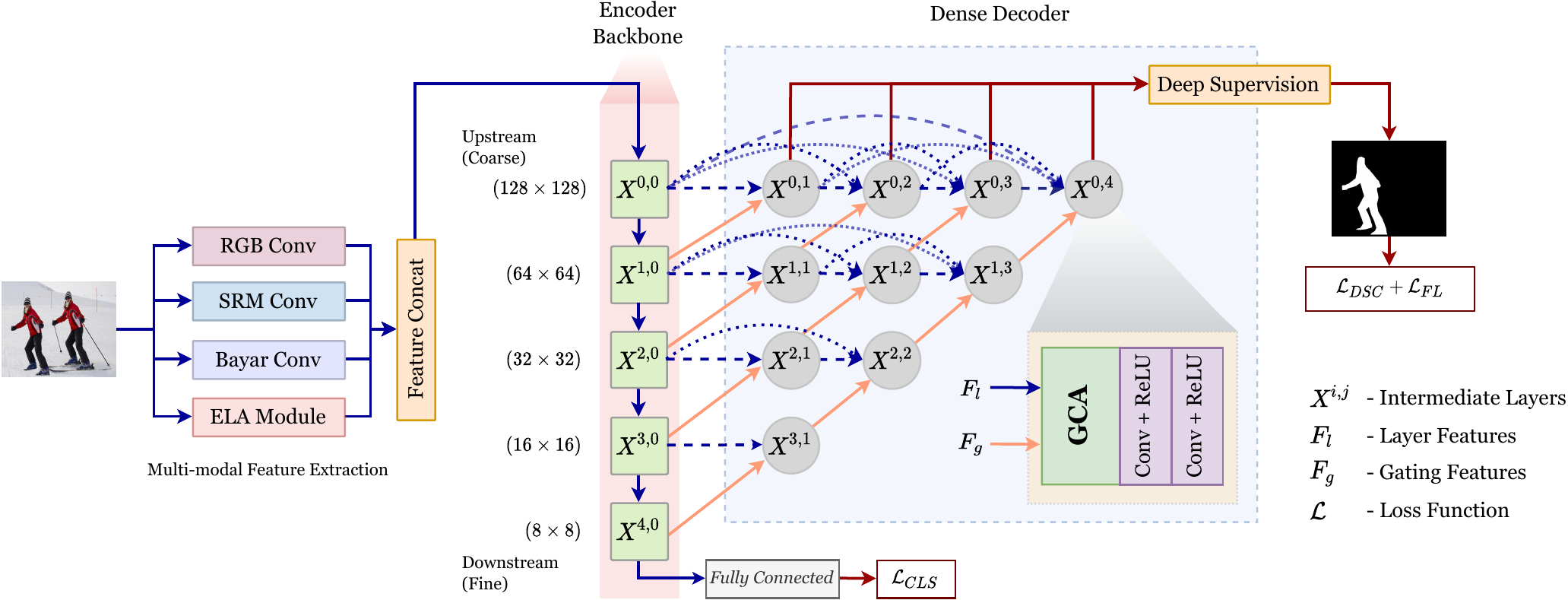}
   \caption{The architecture of GCA-Net. The input image first passes through feature suppression layers similar to \cite{mantranet,span} that extracts multi-modal compression and noise level features. These are then propagated through the EfficientNet encoder backbone consisting of five stages $X^{0,0} \rightarrow X^{4,0}$. Each node of the dense decoder $X^{i,j}$ contains the Gated Context Attention block that takes in the $i^{th}$ layer features $F_l = [X^{i,0}, .. ,X^{i,j-1}]$ shown using \textcolor{blue}{blue} solid and dashed arrows, and the gating feature $F_g$ from $(i+1)^{th}$ layer shown using \textcolor{orange}{orange} arrows. Due to the bottom-up architecture of the decoder, the outputs from zero$^{th}$ layer $X^{0,1\rightarrow4}$ are taken and using deep supervision \cite{unet++} the final localization map is generated. We use separate classification and localization losses for multi-task learning and optimization. }
\label{fig:network}
\end{figure*}

\section{Related Works}

Depending on the type of forgery, different clues and artifacts can be found in an image to determine whether it is authentic or not. These include, compression noise \cite{jpeg_noise}, PRNU sensor information \cite{prnu1, prnu2}, camera model information \cite{camera1, dense_camera}, local noise features \cite{noise1, noiseprint}, etc. Earlier forensic methods used algorithms to detect these noise and signal properties using carefully designed filters. Among them, the SRM filters used in \cite{rich_features} have shown success in identifying hidden noise patterns from images. 

A  number of neural networks have also been proposed for both independent forgery detection \cite{improved_jpeg, detect1, siamese, detetct4}, and localization \cite{local1, camera1, busternet}. Object detection networks using RCNN, region proposal modules, and bounding box identification have shown to be effective for manipulation localization \cite{rich_features, maskrcnn, rcnn2}. Consequently, segmentation networks like Unet and DeepLab have also been used \cite{rru, chen2017deeplab}.
Furthermore, the use of constrained layers have also shown effectiveness in improving input representation and minimizing the effects of dominant spatial features \cite{constrained, yang2020constrained}. ManTra-Net \cite{mantranet} uses a simple VGG \cite{vgg} network with a Z-pooling method to localize anomalous features. This work was extended by SPAN \cite{span} to further model the spatial correlation via local self-attention blocks and hierarchical pyramid propagation. However, both networks fail to utilize the correlation of global context and multi-scale features. In a recent work \cite{pscc}, PSCC-Net was proposed that tries to address these problems using channel correlation. They showed that multi-scales attention could be leveraged to improve manipulation detection.

While self-attention was originally introduced for language modeling \cite{attenionisall}, this seminal work has been proven to improve long-range feature representation across a variety of domains. Several works have utilized variations of attention for forgery detection \cite{ontheface, skip-connection, multi-scale}. Vision tasks primarily use two types of attention mechanisms: channel-attention \cite{squeezeandexcitation} and spatial-attention \cite{non-local}. Spatial attention using non-local blocks, also termed dot-product attention, can be considered a generic framework for non-linear correlation. However, this method suffers from quadratic complexities as the attention maps are computed for every pixel pair. An improved context attention was introduced in \cite{gcnet} that replaces the quadratic operation by simplifying the context modeling framework. In our work, we use this global context framework in conjunction with a  gated attention mechanism to explore discrepancies in spatial and feature channels concurrently. Attention gating has previously been used for various tasks including graph learning \cite{graph-gate}, language processing \cite{nlp-gate}, medical image segmentation \cite{attention-unet}, \textit{etc.} We introduce gating for forgery detection in order to reinforce and filter coarse-level features during the upscaling and decoding phases. This leads to improvements in information sharing by only propagating the necessary forensic signals through the network.


\section{Proposed Method}


\subsection{Overview}

As illustrated in Fig. \ref{fig:network}, GCA-Net is a multi-branch dense encoder-decoder network. The model is comprised of three parts: a feature encoder, a dense decoder, and two heads for classification and localization. An input image first passes through a series of content suppression layers that generate multi-domain semantic features. While standalone CNN layers are capable of extracting features from images, they are localized and biased towards the dominant spatial features rather than the manipulation traces. So, in addition to the constrained layers in \cite{mantranet}, we use an Error Level Analysis (ELA) module to extract signal noise and compression artifacts.
The encoded backbone features and intermediate layer outputs are used by the attention decoder to produce the final localization map. Each decoder block is composed of a GCA layer followed by a series of convolution layers for feature accumulation and upscaling. Additionally, the encoder features also pass through a fully connected layer to generate an image-level probability score. The simultaneous classification and localization utilizes a multi-task learning approach for better feature representation. This plays a significant role in reducing false positives and improving generalizability. Additional details regarding the implementation of ELA, and the effect of each layer on the network's performance can be found in Suppl. A1.




\subsection{Dense Feature Decoder}

For the decoder branch, we use the dense Unet++ \cite{unet++} architecture that utilizes a multi-scale bottom-up architecture for feature propagation. We denote the upper layers of the decoder as \textit{coarse} layers, and the lower layers as \textit{fine} based on the selectivity of their feature maps. The skip connections between the layers allow the flow of multi-modal semantic features, thus improving gradient flow and minimizing feature loss. Moreover, since the decoder samples the intermediate layers at multiple scales, it improves global context representations.

Features from an input $I\in\mathbb{R} ^{3\times H\times W}$ are sampled at five intermediate scales $H/k\times W/k, $ where $k\in \{2, 4, 8, 16, 32\}$. Each \textit{gray} circle in Fig. \ref{fig:network} represents a decoder node denoted $X^{i,j}$, where $i$ indexes the sequential down-sampling layers of the encoder and $j$ indexes a decoder block along the skip pathway at the $i^{th}$ layer. The output $y^{i,j}$ at any node is computed as,
\begin{align}
y^{i, j} &= \mathcal{C}\left(\vartheta\right) \\
\vartheta &= \Theta\left(\Omega\left(\left[y^{i, k}\right]_{k=0}^{j-1}\right), \mathcal{U}\left(y^{i+1, j-1}\right)\right)
\label{eq:gca}
\end{align}
\noindent where $\mathcal{C}(\cdot)$ is a series of convolution operations followed by a ReLU activation, $\Theta (\cdot)$ is the GCA layer, $\Omega(\cdot)$ is a non-local context block, $\mathcal{U}(\cdot)$ denotes an up-sampling layer, and $[\cdot]$ denotes the concatenation layer. 

The bottom-up architecture ensures that at each node features from all lower scales are accumulated. For instance, $X^{3,1}$ is reinforced by $X^{4,0}$ and $X^{3,0}$. Similarly, computation of the final node $X^{0,4}$ aggregates features from all previous layers and nodes $X^{4,0}$ upto $X^{0,3}$. Thus, each node can determine which features are most relevant reinforced by the finer lower-level features and propagate it to upper branches. In contrast to standard encoder-decoder networks, this multi-scale fusion enables easier identification of the essential attributes without relying on a single previous output.

\subsection{Gated Context Attention (GCA)}

\begin{figure}
\centering
\includegraphics[width=0.98\linewidth]{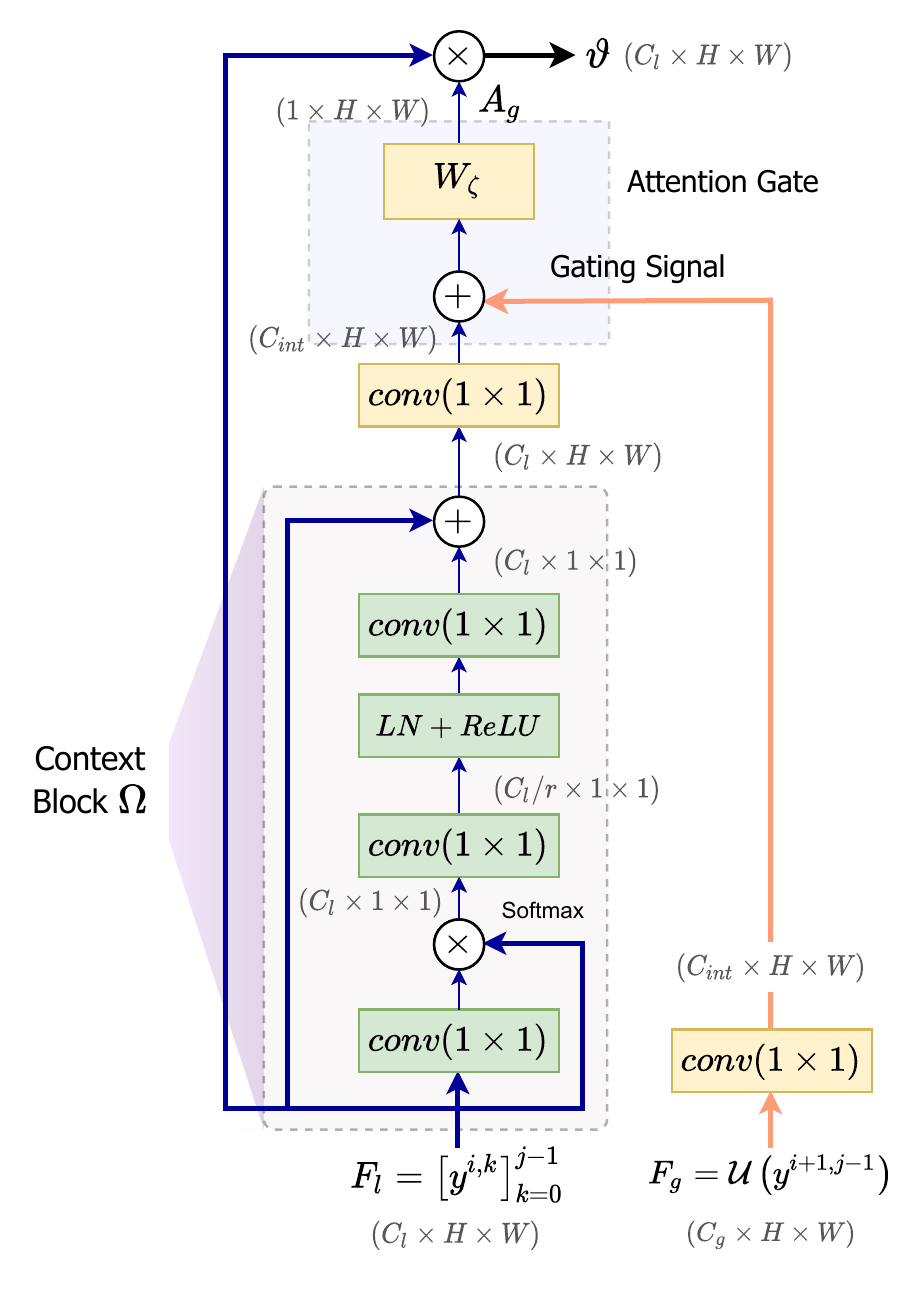}
\caption{The structure of Gated Context Attention block.
The \textcolor{blue}{blue} line shows the flow concatenated features $F_l$ of $i^{th}$ decoder level and the \textcolor{orange}{orange} line denotes the upsampled gating feature $F_g$ from $(i+1)^{th}$ level as referred in Fig. \ref{fig:network} }
\label{fig:gca}
\end{figure}

Attention mechanisms are used to modulate learned features according to their relative significance. The GCA operation shown in Fig. \ref{fig:gca} is divided into two primary stages: 1) Global Context Pooling and 2) Attention Gating.

Deep convolution stacks tend to obfuscate global pixel-to-pixel relationships due to their nature of locality \cite{non-local}. Non-local blocks solve this problem using attention weights and aggregating information from other points to reinforce the features of a query position. Global context modeling \cite{gcnet} is an improved attention framework for identifying long range dependencies between feature maps. In the first stage, we compute the global context from the concatenated features of the current level. For instance, to compute the context of node $X^{0,3}$, we aggregate $[X^{0,0}, X^{0,1}, X^{0,2}]$ which is denoted as $F_l$ or \textit{layer features}. These are coarse feature maps containing a greater amount of global information than the subsequent finer layers. Because identification of manipulation features is based on detecting changes between a group of pixels and their surroundings, using these global contexts allows the model to recognize the differences between altered regions. We can rewrite Eq. (\ref{eq:gca}) as, $\vartheta = \Theta\left(\Omega\left(F_l\right), F_g\right)$,
where, $F_l = \left[y^{i, k}\right]_{k=0}^{j-1}$ is the concatenation of $i^{th}$ level features, and $F_g = \mathcal{U}\left(y^{i+1, j-1}\right)$ is the up-sampled feature of $(i+1)^{th}$ level. 

The context block $\Omega$ has three steps $-$ i) attention pooling, ii) feature transform, and iii) feature fusion. For the pooling step we take $F_l \in \mathbb{R} ^{C_l\times H\times W}$ and pass it through a $1\times1$ convolution to reduce the channels to $C_l\times1\times1$. It groups the features of all positions via weighted averaging to obtain the global context vector. This is similar to the Global Average Pooling of Squeeze-Excitation (SE) layer \cite{squeezeandexcitation}. The pooled vector is then passed through a bottleneck block with a factor $r$ to capture the channel-wise dependencies. This is the transform step. The reduction and expansion is analogous to the excitation operation of SE block. Finally, the fusion step aggregates the context features to the features of each input position using a broadcast element-wise addition.

The second stage of GCA is the attention gating which is used to filter the coarse level features. Attention Gates (AG) identify salient image regions and prune feature responses to retain only relevant activations. As a result, signals from irrelevant background regions and spatial contexts are gradually suppressed. AGs generate a coefficient matrix, $A_g \in [0,1]$, which is multiplied with the input to alter the scale of the activations. Each AG learns to focus on a subset of target structures reinforced by the downstream features. The coarse maps of the encoder represent global relationships, while the downstream layers identify \textit{finer} discriminating features. Gating uses these finer embeddings to disambiguate irrelevant and noisy responses within the coarse features. At each decoder node $X^{i,j}$, the gating feature $F_g$ is the upsampled feature propagated by the node $X^{i+1,j-1}$, highlighted in \textit{orange}. In order to obtain the gating coefficient via additive attention, we first transform the context feature $\Omega\left(F_l\right) \in \mathbb{R} ^{C_l\times H\times W}$ and gating feature $F_g \in \mathbb{R} ^{C_g\times H\times W}$ to an intermediate vector $F_{int} \in \mathbb{R} ^{C_{int}\times H\times W}$. A non-linear transformation layer $W_\zeta = ReLU(Conv(\cdot))$ is then used to resample this vector. This produces the final attention matrix $A_g = W_\zeta(\Omega\left(F_l\right) \oplus F_G) \in \mathbb{R} ^{1\times H\times W}$. Finally, the attention matrix is multiplied with the input feature stack $F_l$ to generate $\vartheta \in \mathbb{R} ^{C_l\times H\times W}$, the GCA output. The resulting gated features are sent through a series of convolution and activation layers $\mathcal{C(\cdot)}$ that perform the decoding operation at the particular node.
Additional spatial and channel attention layers \cite{scse} can be added after this stage to further modulate the learnt features.

\section{Experiments}

\subsection{Datasets}

We follow the evaluation protocols in \cite{mantranet, rich_features} for training and validation.
We train our model on four types of data $-$ splicing, copy-move, inpainting, and authentic images; collected from Dresden \cite{dresden},  MS COCO synthetic \cite{lstm}, Defacto \cite{defacto}, and IMD-Real \cite{imd2020} datsets. For both pre-trained and finetuned evaluation we use the four standard datasets: CASIAv2 \cite{casia}, NIST16 \cite{nist}, COVERAGE \cite{coverage} and IMD-2020 \cite{imd2020}, following the training/testing split described in \cite{pscc}. In total, we used ${\raise.17ex\hbox{$\scriptstyle\sim$}}170$k images for training, with roughly equal class distribution. This amount is significantly lower compared to existing SOTA methods like SPAN \cite{span}, MantraNet \cite{mantranet}, and PSCC-Net \cite{pscc} that use upwards of $500$k${\raise.17ex\hbox{$\scriptstyle\sim$}}1$M images. We did not use a larger dataset due to resource and accessibility constraints. The majority of our computation was conducted on an Nvidia 1080 Ti GPU. However, despite the smaller train set, our model outperforms these methods in multiple experiments.

\subsection{Loss Function}

The train data consists of input images $I \in \mathbb{R}^{3\times H\times W}$ and binary ground-truth masks $M \in [0,1]^{1\times H\times W}$. GCA-Net was trained using a multi-task loss function combining detection and localization losses. 
Most traditional methods \cite{mantranet, span, pscc} train the localization network using Binary Cross-Entropy (BCE) loss (eq. \ref{eq:bce}). BCE works well for classification problems with lots of data and a balanced dataset because it weighs all predictions equally.

\begin{equation}
\mathcal{L}_{BCE} = -\frac{1}{N} \sum_{i=0}^{N} y_{i} \cdot \log \left(\hat{y}_{i}\right)
\label{eq:bce}
\end{equation}
The loss function minimizes the distance between the predicted and ground truth value for all predictions. For example, if a fake pixel is predicted with a probability of $0.7$, BCE will try to bring it closer to $1$ in order to minimize the loss. Thus, for unbalanced datasets, predictions become skewed toward a particular class. Since we are working with a small and unbalanced dataset, we opt to use a combination of Dice \cite{dice}, and Focal \cite{focal} loss. We compute the dice loss using the following equation,
\begin{equation}
\mathcal{L}_{DSC}(P, G) = -\log \left( \frac{2 \cdot |P \cap G| + \epsilon}{|P|+|G|+\epsilon} \right)
\end{equation}
For a prediction mask $P$ and ground-truth $G$ both having dimensions $1\times H\times W$, the numerator calculates the intersection of the regions, and the denominator measures the union. Dice loss minimizes the distance between the predicted and true regions. Rather than calculating loss for each individual pixel, it improves prediction for the entire region. This is critical in lowering our false positive rate. Since BCE tries to minimize loss for each pixel independently, the network always produces incorrect predictions for a certain number of pixels. 

Even though Dice loss improves false positive and region overlaps, it falls short for small forged regions. This is because as the regions become very small, $|G|$ becomes small. So if the model does not predict anything at all, \textit{i.e.,} if $|P|\to0$, then $|P \cap G|\to0$, and the total loss decreases. Thus, dice loss alone is ineffective in these instances. To overcome this problem, we combine Focal loss with Dice loss. Focal loss is an improvement over BCE loss with an additional temperature parameter to account for overconfident predictions. Focal loss is calculated as,
\begin{equation}
\mathcal{L}_{FL}\left(P\right)=-\alpha_{\mathrm{t}}\left(1-P\right)^{\gamma} \log \left(P\right)
\end{equation}
The temperature $\gamma$ controls the weighting of each prediction. When $\gamma > 0$ is used, weak predictions are weighted more heavily. So, the network focuses on improving a prediction of $0.3$ towards $1$ rather than trying to improve a prediction of $0.7$. Since we are trying to correctly classify fake pixels, a larger prediction $P$ results in a smaller $(1-P)^\gamma$, thereby reducing the overall loss. So, the network will try to classify weaker predictions more accurately. Combining all of the above, our final loss function becomes,
\begin{equation}
    \mathcal{L} = w_c \cdot \mathcal{L}_{CLS} + w_d \cdot \mathcal{L}_{DSC} + w_f \cdot \mathcal{L}_{FL}
\end{equation}
$\mathcal{L}_{CLS}  = \mathcal{L}_{BCE}$ is the classification loss for the encoder's prediction. By combining classification loss, the network can optimize both the encoder and decoder at the same time. Each loss parameter can be adjusted independently. We use $\epsilon = 10^{-7}, \gamma = 2, w_c=1, w_d=1.10,$ and $w_f=1.15$ to train our final network. In Table \ref{tab:loss} we report the results of different configuration of loss functions on the CASIAv2 validation set. To quantify localization performance, we use pixel-level AUC following previous works \cite{mantranet, rich_features, pscc} and Dice score \textit{i.e.}, pixel F1 score.

\begin{table}
\centering
\resizebox{0.9\linewidth}{!}{%
\begin{tabular}{l|c|c} 
\hline
Loss Function               & AUC   & F1     \\ 
\hline
\hline
$\mathcal{L}_{BCE}$                        & 78.33 & 60.29  \\
$\mathcal{L}_{DSC}$                       & 84.12 & 64.41  \\
$\mathcal{L}_{DSC} + \mathcal{L}_{FL}$                & 86.05 & 65.50  \\
$\mathcal{L}_{DSC} + \mathcal{L}_{FL}-Reduction$ \cite{reducedfocal}       & 86.81 & 67.25  \\
$\mathcal{L}_{CLS} + \mathcal{L}_{DSC} + \mathcal{L}_{FL}-Reduction$ & 88.95 & 72.39  \\
\hline
\end{tabular}
}
\caption{Localization performance of GCA-Net under different loss functions on CASIAv2 validation set. \vspace{-8pt}}
\label{tab:loss}
\end{table}

\subsection{Ablation Study}
\label{sec:abl}
We follow the ablation study in \cite{gcnet} to compare changes of parameters and configuration of the network. The following experiments are done on the CASIAv2 validation set, and we report the pixel-level AUC and F1 score. 

\begin{table}[h]
\centering
\resizebox{0.72\linewidth}{!}{%
\begin{tabular}{l|cc} 
\hline
\multicolumn{3}{c}{\rule{0pt}{1em}  (a) Block Design}                   \\ 
\hline \hline
\multicolumn{1}{c|}{} & AUC(\%)       & F1             \\ 
\cline{2-3}
\Tstrut
baseline (without GCA)              & 85.2          & 68.4           \\
non-local             & 86.1          & 69.8           \\
simplified non-local  & 86.9          & 70.5           \\
global context        & 87.4 & 71.7  \\ 
GCA                 & \textbf{88.9} & \textbf{72.4}  \\ 
\hline
\multicolumn{3}{c}{\rule{0pt}{1em} (b) Bottleneck Design}              \\ 
\hline \hline
\Tstrut
baseline (without GCA)              & 85.2          & 68.4           \\
w/o ratio             & \textbf{88.9} & \textbf{72.5}  \\
r16 (ratio 16)        & 87.3          & 70.4           \\
r16+ReLU              & 87.3          & 70.4           \\
r16+LN+ReLU           & 88.3          & \textbf{72.6}  \\ 
\hline
\multicolumn{3}{c}{\rule{0pt}{1em} (c) Bottleneck Ratio}               \\ 
\hline \hline
\Tstrut
baseline (without GCA)              & 85.2          & 68.4           \\
ratio 4               & \textbf{88.8} & \textbf{72.8}  \\
ratio 8               & 88.5          & 72.8           \\
ratio 16              & 88.3          & 72.6           \\
ratio 32              & 87.9          & 71.7           \\ 
\hline
\multicolumn{3}{c}{\rule{0pt}{1em} (e) Block Positions}                \\ 
\hline \hline
\Tstrut
baseline (without GCA)              & 85.2          & 68.4           \\
all decoder           & \textbf{88.3} & \textbf{72.6}  \\
only end nodes        & 87.1          & 70.5           \\
only intermediates    & 87.8          & 71.9           \\
only top nodes        & 86.3          & 69.4           \\
\hline
\end{tabular}
}
\caption{Ablation study of the GCA block for pixel-level localization on the CASIAv2 validation set. }
\label{tab:abl}
\end{table}

\begin{description}[leftmargin=*]
\vspace{3pt}
    \item[Block Design:] For the choice of long range context modeling we compare other existing frameworks - Non-local (NL) block, Simplified Non-local (SNL) block, SE block, and the Global Context block, placed before every decoder node. Table  \ref{tab:abl}(a) shows all context frameworks achieve better performance over baseline. NL and SNL blocks are quite similar, while GCA blocks with comparably fewer parameters  yield the best performance.
    \vspace{4pt}
    \item[Bottleneck Design:] The effects of each component in the bottleneck section are shown in Table \ref{tab:abl}(b). w/o ratio uses a single $1 \times 1$ convolution as a transform which has higher parameters and achieves the best performance. Even though r16+ReLU has fewer parameters they are harder to optimize. Thus Layer Norm (LN) is used to ease optimization, leading to performance similar to w/o ratio.
    \vspace{4pt}
    \item[Bottleneck Ratio:] The bottleneck is used to reduce redundancy in parameters and provide a trade-off between parameter and performance. The bottleneck ratio $r$ controls the amount of feature compression. Table \ref{tab:abl}(c) shows that the network's performance improves consistently as the ratio decreases. We use a bottleneck ratio $r=4$ which has a good balance of performance and parameters. 
    \vspace{4pt}
    \vspace{4pt}
    \item[Block Positions:] We determine whether the placement of the GCA blocks have an effect on performance. Various placement positions are illustrated in Suppl. A2. As seen in Table \ref{tab:abl}(e), this effects performance only to a small degree. The best results are obtained by placing the attention block before each node.
\end{description}
\vspace{3pt}
Additional ablation experiments regarding backbone choice, and training specifics are provided in Suppl. A3.

\section{Comparison and Evaluation} 

We compare GCA-Net against existing SOTA architectures for both pre-trained and finetuned evaluation. The pre-trained model was selected based on the best validation score on the train set. We report existing values as mentioned in \cite{pscc}. For finetuned evaluation we use unseen test splits generated following the evaluation process in \cite{rich_features}.

\begin{figure*}
\centering
\includegraphics[width=\linewidth]{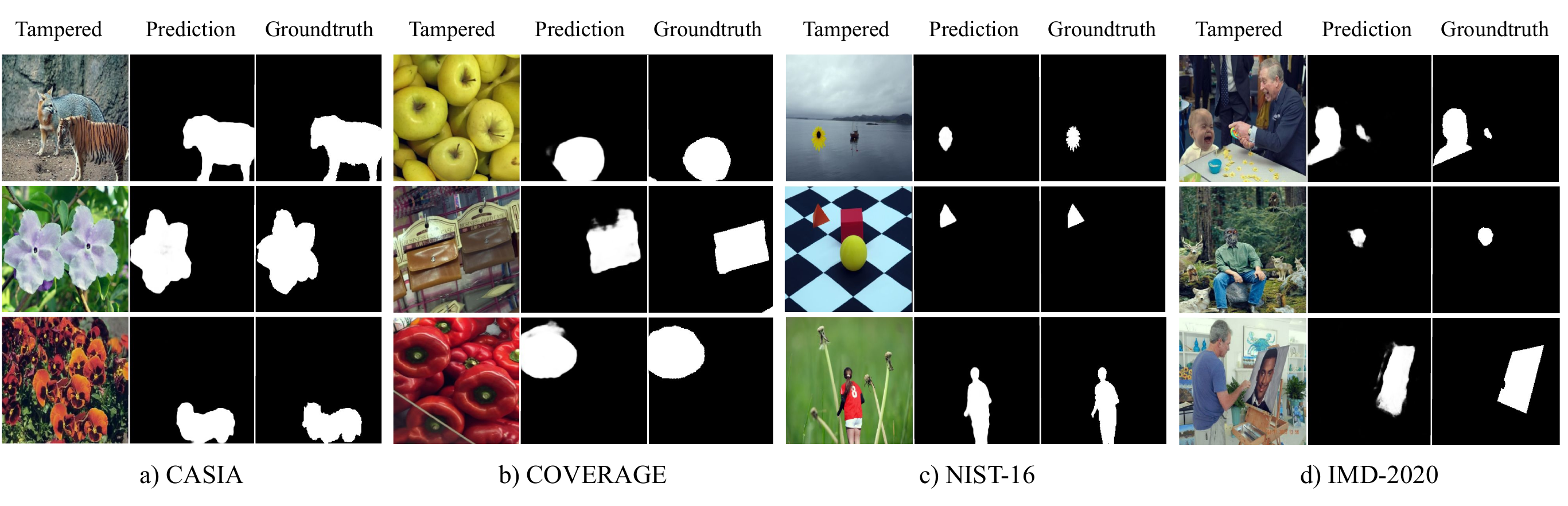}
   \caption{Qualitative examples of manipulation localization with their respective groundtruths from multiple datasets. }
\label{fig:results}
\end{figure*}

In Tables \ref{tab:pre}, \ref{tab:res} we compare GCA-Net to existing methods. Both the pre-trained and fine-tuned comparisons demonstrate that GCA-Net outperforms all other methods on the CASIAv2 and IMD-2020 datasets and is comparable on the NIST and COVERAGE datasets. On CASIA and IMD, we see an improvement of $5.4\%$ and $4.46\%$ respectively over PSCC-Net. GCA-Net achieves an AUC of $95.3$ on NIST-16, trailing behind PSCC-Net by only $4\%$. However, we surpass every network on NIST-16 in terms of F1 score. This is because of our fine-tuning the loss functions. Pixel-level F1 score measures the region overlap of the prediction and ground-truth. Since we optimized the network using Dice loss, its region identification is better than the existing models. We rank third in COVERAGE, behind SPAN and PSCC-Net. COVERAGE includes samples with very small shifts of copied regions followed by contrast correction and edge blurring. Our train data, which is composed entirely of publicly available datasets, is free of such perturbations, resulting in a difference in the train-test distribution. This limitation can be overcome by training on synthetic copy-move data supplemented with adversarial examples.
Qualitative examples of localization are shown in Fig. \ref{fig:results}. Additional examples comparing localization for both authentic and forged images against MantraNet is provided in the supplementary.

\begin{table}
\centering
\resizebox{\linewidth}{!}{%
\begin{tabular}{l|cccc} 
\hline
Method    & CASIAv2       & COVERAGE      & NIST16        & IMD2020        \\ 
\hline
\hline
MantraNet \cite{mantranet} & 81.7          & 81.9          & 79.5          & 74.8           \\
SPAN \cite{span}     & 79.7          & \textbf{92.2} & 84.0          & 75.0           \\
PSCC-Net \cite{pscc}  & 82.9          & 84.7          & \textbf{85.5} & 80.6           \\ 
\hline
GCA-Net   & \textbf{87.1} & 83.1          & 85.2          & \textbf{81.3}  \\
\hline
\end{tabular}
}
\caption{Comparison of localization AUC against existing methods using their pre-trained models.}
\label{tab:pre}
\end{table}

\begin{table}
\centering
\resizebox{\linewidth}{!}{%
\begin{tabular}{l|llll} 
\hline
Methods        & CASIAv2              & COVERAGE    & NIST16               & IMD2020               \\ 
\hline
\hline
ELA \cite{ELA2}           & 61.3 / 21.4          & 58.3 / 22.2 & 42.9 / 23.6          & -                     \\
NOI \cite{wavelet}           & 61.2 / 26.3          & 58.7 / 26.9 & 48.7 / 28.5          & 58.6 / -              \\
CFA1 \cite{cfai}           & 52.2 / 20.7          & 48.5 / 19.0 & 50.1 / 17.4          & 48.7 / -              \\ 
\hline
RGB-N \cite{rich_features}         & 79.5 / 40.8          & 81.7 / 43.7 & 93.7 / 72.2          & -                     \\
SPAN \cite{span}          & 83.8 / 38.2          & 93.7 / 55.8 & 83.6 / 29.0          & 75.0 / -              \\
PSCC-Net \cite{pscc}      & 87.5 / 55.4          & \textbf{94.1 / 72.3} & \textbf{99.6} / 81.9 & 80.6 / -              \\ 
\hline
GCA-Net & \textbf{92.2 / 71.2} &     87.4 / 69.5        & 95.3 / \textbf{84.5} & \textbf{86.4 / 42.6}  \\
\hline
\end{tabular}
}
\caption{Evaluation against existing fine-tuned models for pixel-level localization AUC/F1 score on unseen test splits.}
\label{tab:res}
\end{table}

\subsection{Detection Performance}

To analyze the image-level detection performance, we compare pretrained GCA-Net to SOTA architectures using the metrics reported in \cite{pscc}. We use a detection dataset containing $511$ forged and $749$ real images taken from CASIAv2. As can be seen from Table \ref{tab:detc}, GCA-Net significantly outperforms all other models. This is because it was equipped with a dedicated classification head that was trained particularly to recognize images with forged content. In comparison, existing approaches identify the image class by counting the number of manipulated pixels discovered in the localization result.

\begin{table}[h]
\centering
\resizebox{0.8\linewidth}{!}{%
\begin{tabular}{l|c}
\hline
Method               & \multicolumn{1}{l}{\rule{0pt}{1em} Image-Level F1 Score}  \\ 
\hline \hline
\rule{0pt}{1em}MantraNet            & 56.69                                     \\
SPAN                 & 63.48                                     \\
PSCC-Net             & 66.88                                     \\
\hline
GCA-Net (pretrained) & \textbf{85.51}                                     \\
\hline
\end{tabular}
}
\caption{Comparison of image-level detection performance on CASIAv2 detection set against other methods. \vspace{-10pt}}
\label{tab:detc}
\end{table}

\subsection{False Positive Evaluation}

One of the primary contributions of our proposed method is the reduction of false positives in localization. In order to evaluate the degree of improvement against other networks, we calculate \textit{FPR} or \textit{False Positive Rate} for both authentic and manipulated images. \textit{FPR} quantifies the proportion of data in a sample that is incorrectly identified. In our case, a false positive prediction occurs when a pixel that should be authentic, i.e., $0$, is classified as fake, i.e., $1$. \textit{FPR} measures the fraction of incorrect predictions made against the entire set of pixels in an image $FPR=\frac{False ~Positive ~Pixels}{Total ~Number ~of ~Pixels}$.
However, because the number of incorrect predictions is relatively small in comparison to the total number of pixels in an image, this value can become extremely small and difficult to interpret. Thus, we use the $-log(FPR)$ as our evaluation metric. Increased values indicate better performance and lower false positives.

\begin{table}[h]
\centering
\resizebox{0.98\linewidth}{!}{%
\begin{tabular}{cl|c|c} 
\hline
                                 &                & \multicolumn{1}{l|}{ManTraNet} & \multicolumn{1}{l}{GCA-Net}  \\ 
\hline
\hline
\Tstrut
\multirow{3}{*}{CASIA} & Authentic (A)  & 5.39                           & 9.88                         \\
                                 & Tampered (T)   & 4.77                           & 5.09                         \\
                                 & Combined (A+T) & 4.95                           & 5.97                         \\ 
\hline
\hline
\Tstrut
\multirow{3}{*}{IMD}   & Authentic (A)  & 4.57                           & 6.10                         \\
                                 & Tampered (T)   & 4.43                           & 5.65                         \\
                                 & Combined (A+T) & 4.66                           & 6.93                         \\ 
\hline
\hline
\multicolumn{2}{l|}{Authentic (CASIA + IMD)}      & 4.94                           & 8.83                         \\
\multicolumn{2}{l|}{Tampered (CASIA + IMD)}       & 4.56                           & 5.30                         \\
\hline
\end{tabular}
}
\caption{Comparison of false positive, $-log(FPR)$, for authentic and manipulated images from CASIAv2 and IMD2020 test sets. Higher values denote less false positives. The experiments were done separately on authentic and tampered images, as well as a combination of both from the two datasets.}
\label{tab:fpr}
\end{table}

To conduct the evaluation, we created a test set using $200$ authentic and $100$ tampered images taken independently from both CASIAv2 and IMD2020 datasets. We used the pretrained model for GCA-Net, and the publicly available implementation for ManTraNet.  As illustrated in Table \ref{tab:fpr}, GCA-Net consistently shows better performance. For authentic images in both CASIA and IMD, GCA-Net's score is almost double that of the other methods. Moreover, due to the logarithmic nature of the metric, fairly small differences denote significant real world improvement. Additional qualitative samples of authentic predictions are provided in the supplementary.

\subsection{Robustness Analysis}

\begin{figure}
\centering
\includegraphics[width=\columnwidth]{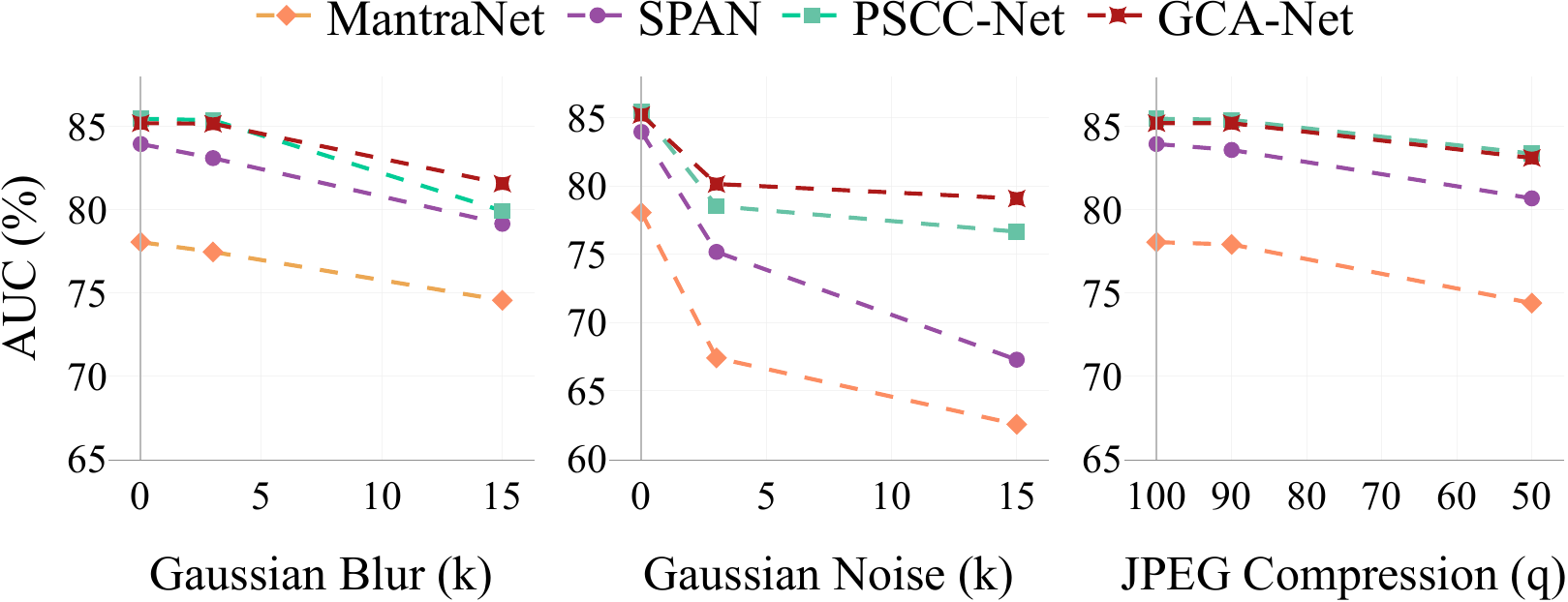}
   \caption{Comparison of model robustness against different post-processing methods. }
\label{fig:robust}
\end{figure}

We examine the performance of our proposed method against various attacks/post-processing to further verify its efficacy and robustness. For this purpose, we degrade images from the NIST16 test set using the distortion settings in \cite{pscc}. These include Gaussian Blur with kernel size \textit{k}, JPEG Compression with a quality factor \textit{q}, and Additive Gaussian Noise using standard deviation $\sigma$. The reported metrics are calculated using the pretrained GCA-Net. In Fig. \ref{fig:robust}, we can see that our model outperforms existing approaches against various post-processing attacks.

\subsection{Limitations}

\begin{figure}
\includegraphics[width=\linewidth]{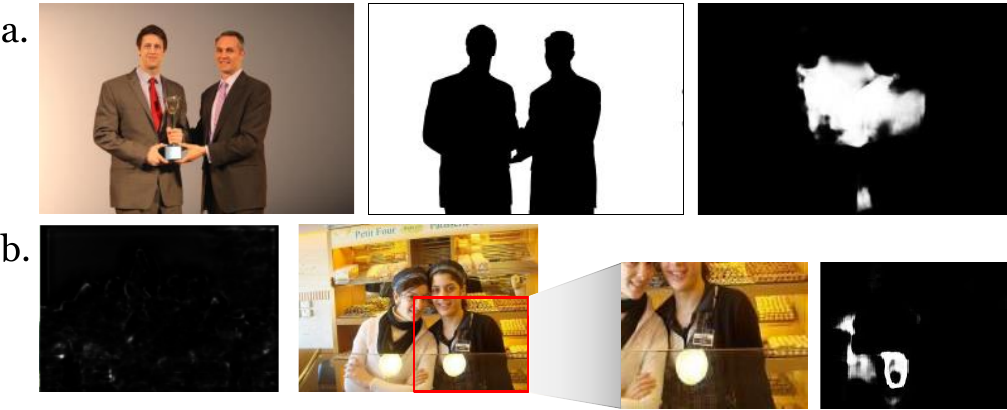}
   \caption{(a) Example of a failed prediction. The groundtruth (2nd image) shows white pixels as forged and black as authentic. The prediction (3rd image) identified discrepancies between the two regions, but inverted the classes. (b) The left pair shows example of a failed localization for a large resolution image. In the right pair, for the same image when tested with a cropped portion, the network successfully located the forged pixels.}
\label{fig:limitation}
\end{figure}

In our experiments we found that GCA-Net might fail when the manipulation region is very large compared to authentic pixels. Fig. \ref{fig:limitation}(a) contains a  sample from NIST-16 where the entire white region is manipulated. Although GCA-Net could detect the discrepancy between the authentic and forged regions, it is not confident about the prediction. It highlighted the centre portion denoting that the region is different from the surrounding pixels. For a second example, we tested our network similar to \cite{mantranet} to check for manual assistance applicability. The initial image in Fig. \ref{fig:limitation}(b) was of dimension $1024\times1520$ having a small forged region. The network failed to locate the region when tested with the entire image. After that, we cropped the image around the forged location and again tested the cropped image. This time the network was able to identify the manipulated region. This indicates that GCA-Net could be used as a computer-aided tool.

\section{Conclusion}

In this paper, we introduced a novel Gated Context Attention Network (GCA-Net) for detecting and localizing image forgeries.  Our proposed network uses a gated attention block to utilize the global context features together with the region attributes to localize manipulated pixels. The proposed attention framework improves long-range dependency modeling and reduces attenuation of the hidden forensic features. This paper illustrated the problems surrounding existing methods and how they might be addressed with better feature representation and training strategies. As demonstrated by our results, GCA-Net outperforms existing SOTA architectures on multiple benchmark datasets by upto $6\%$ with significantly lower false positives. In the future, we will further improve the method in handling large resolution images and work on reducing the limitations. We will also explore the model's viability in detecting deep learning-based deepfake forgeries and other semantic segmentation tasks.

\twocolumn[{%
 \centering
 \Large \textbf{Supplementary Appendix \vspace{50pt}}\\
}]

\section*{A1. Content Suppression Modules}
\label{sec:suppression}

\begin{figure}[!h]
\centering
\includegraphics[width=\columnwidth]{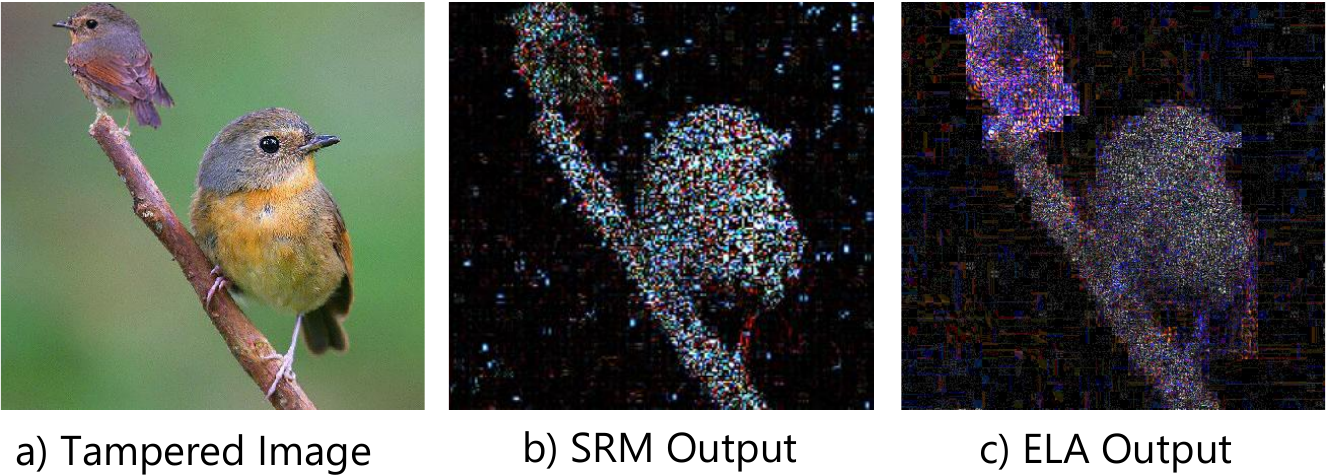}
\caption{Result of SRM filters and ELA on a tampered image. }
\label{fig:ela}
\end{figure}

In order to improve detection of the subtle forensic features and suppress the spatial content of the image, we add additional modules to the encoder's first layer that extract noise level features. For this purpose, we introduced four modules  $-$ i) the \texttt{SRMConv} \cite{rich_features} layer, ii) the \texttt{BayarConv} \cite{constrained} layer, iii) the classic convolution layer termed as \texttt{RGBConv}, and iv) our proposed Error Level Analysis (ELA) Module. Fig. \ref{fig:ela} shows the output of applying SRM and ELA on a tampered image.

ELA has previously been used for localizing compression artefacts from JPEG images \cite{ela}. It works by comparing the pixel-wise difference between an image and its compressed copy. If an image contains pixels from a different source, then the pixels of the two sources would produce different levels of compression noise. We propose to use this ELA output as a feature for the encoder. We take an input image and compress it with a reduced $90\%$ compression factor. Then we calculate the difference between the original and the compressed image to generate the ELA output. This output ELA image is then passed through a series of convolution layers before applying activation to produce the ELA feature map. 

To evaluate the effect of these modules on the encoder, we compare the detection accuracy on the CASIAv2 validation set in Table \ref{tab:conv}. We can see that the choice of the first layer affects model performance to a large amount. The proposed ELA module has a notable effect as it improves encoder accuracy by a factor of more than $3\%$. So, for our final encoder, we select a combination of the four layers. The input images pass through all of them simultaneously, then the outputs are concatenated and sent to the backbone. This additional compression and steganalysis feature helps the network to detect the traces of the boundary regions. Moreover, the encoder becomes more robust to post-processing operations as it learns to detect and correlate the multi-domain artifacts with other spatial features.

\begin{table}[h]
\centering
\resizebox{0.96\linewidth}{!}{%
\begin{tabular}{l|l|c} 
\hline
\Tstrut
1st Conv Layer & \#Filters, Kernel Size & Encoder Accuracy (\%)  \\ 
\hline
\hline
\Tstrut
\texttt{RGBConv}     & 16, k=(3,3), p=1, x2             & 84.61               \\
\texttt{SRMConv}     & 3, k=(5,5)              & 86.77               \\
\texttt{BayarConv}   & 3, k=(5,5)             & 85.25               \\
ELA Module     & 32, k=(3,3), p=1, x2             & 87.03               \\
Combined       & 54, -            & \textbf{88.25}      \\
\hline
\end{tabular}
}
\caption{Results of using additional feature extraction layers for the 1st encoder layer with an EfficientNet-B4 backbone. The results compare only the encoder detection accuracy for an image-level binary classification test on the CASIAv2 validation set.}
\label{tab:conv}
\end{table}

\section*{A2. Additional Ablation Experiments}

\subsection*{A2.1 Block Positions}

\begin{figure}[h]
        \centering
        \begin{subfigure}[b]{0.475\linewidth}
            \centering
            \includegraphics[width=\textwidth]{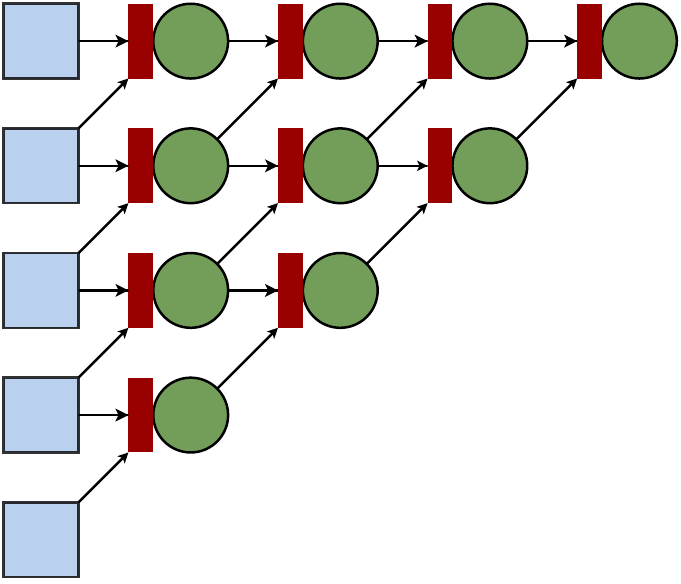}
            \caption{{\small All decoders}}%
        \end{subfigure}
        \hfill
        \begin{subfigure}[b]{0.475\linewidth}  
            \centering 
            \includegraphics[width=\textwidth]{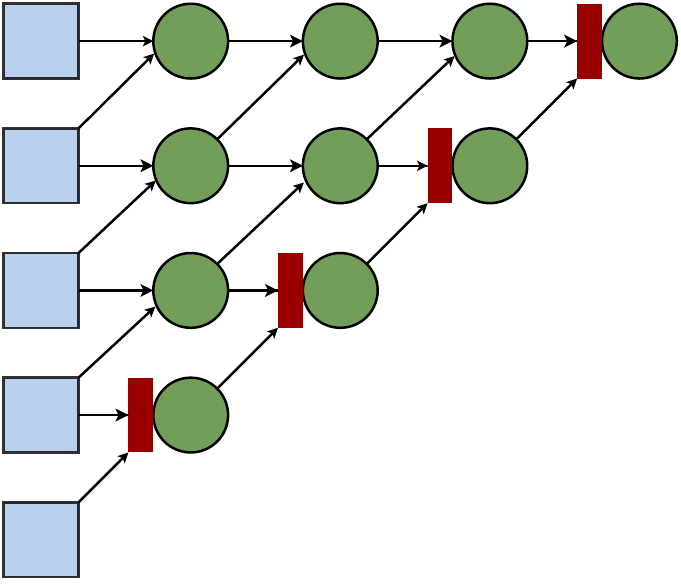}
            \caption{{\small Only end nodes}}    
        \end{subfigure}
        \vskip\baselineskip
        \begin{subfigure}[b]{0.475\linewidth}   
            \centering 
            \includegraphics[width=\textwidth]{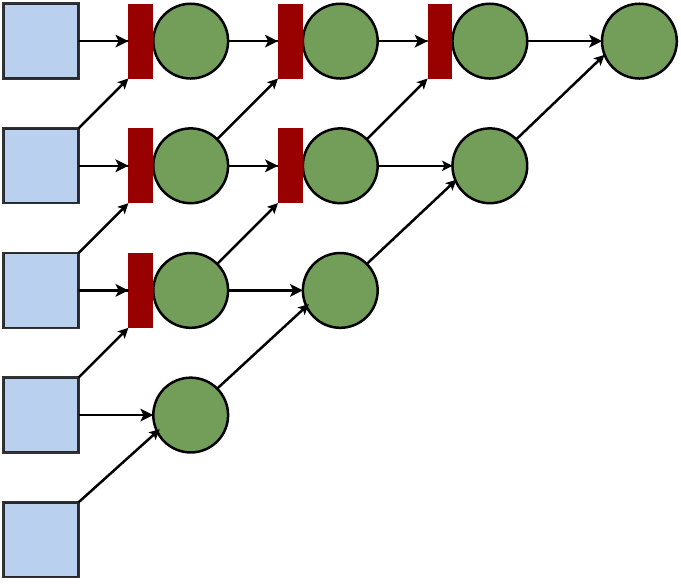}
            \caption[]{{\small Only intermediate nodes}}    
        \end{subfigure}
        \hfill
        \begin{subfigure}[b]{0.475\linewidth}   
            \centering 
            \includegraphics[width=\textwidth]{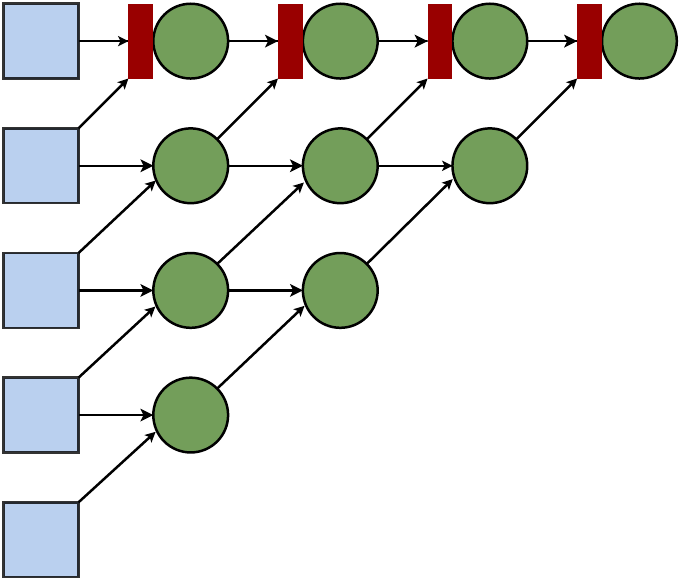}
            \caption[]{{\small Only top nodes}}    
        \end{subfigure}
        \caption{ Different positions of placing the GCA block. Blue squares are encoder layers, Green circles are decoder nodes, and Red rectangles represent the GCA block.} 
        \label{fig:posns}
\end{figure}

In Section 4.3 we had talked about the effects of placing the GCA block in different positions within the network. Fig. \ref{fig:posns} shows these placement positions. The blue squares represent the encoder layer, green circles are the decoder nodes, and the red rectangles denote the GCA block. 

\begin{figure*}
\centering
\includegraphics[width=0.96\linewidth]{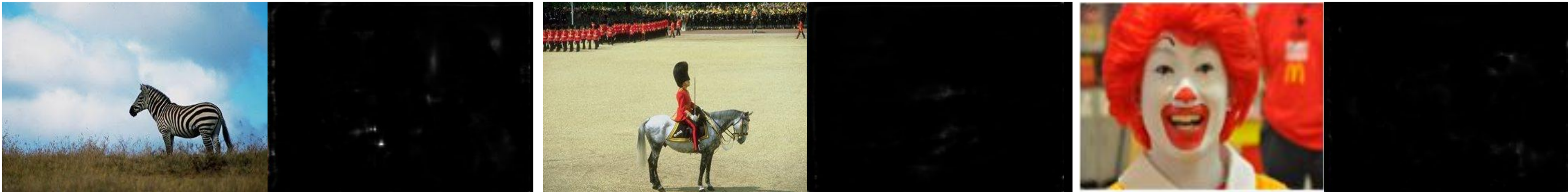}
  \caption{Localization for the three authentic images previously shown in Fig. 2 of the paper. Since groundtruth masks for pristine images are blank, they are not shown here. GCA-Net predicts almost blank masks for authentic images with minimum false positives.}
\label{fig:auth_res}
\end{figure*}

\subsection*{A2.2 Backbone Choice}

There are no dominant network architectures proven to be useful for IFLD tasks. XceptionNet has been shown to perform well for DeepFake detection, and media forgeries \cite{faceforensics}. DenseNet also showed promise in detecting camera model features \cite{dense_camera}, which has relevant implications for manipulation identification. We test multiple such backbone networks to test their efficacy for manipulation detection. We trained and tested these baseline models using the CASIAv2 \cite{casia} dataset. Since we are evaluating the encoder performance only, we perform these tests as a classification task without the decoder and compare the image-level detection performance. From Table \ref{tab:cnn} we see that EfficientNet performs the best. 
Additionally, it uses an inverse bottleneck convolution with channel attention making it the lightest of all the networks with only $19.34$ million parameters. 

\begin{table}[t]
\centering
\resizebox{0.95\linewidth}{!}{%
\begin{tabular}{l|c|c} 
\hline
\Tstrut
Model           & \#Params (M) & Encoder Accuracy (\%)  \\ 
\hline
\hline
\Tstrut
XceptionNet \cite{chollet2017xception}    & 22.86    & 78.03                          \\
DenseNet-161 \cite{densenet}   & 28.68    & 83.56                          \\
ResNeXt-50 \cite{resnext}     & 30.42    & 82.29                          \\
SEResNeXt-50 \cite{squeezeandexcitation}   & 27.56    & 85.81                          \\
EfficientNet-B4 \cite{tan2020efficientnet} & 19.34    & \textbf{87.65}                 \\
\hline
\end{tabular}
}
\caption{Baseline detection accuracy of different architectures for image-level binary classification on CASIA validation set.}
\label{tab:cnn}
\end{table}

\section*{A3. Implementation Details}

In order to tackle the challenge of low data and improve generalizability, all images were augmented using Flipping, Random Rotations, Optical and Grid Distortions, and Gaussian Blur, each with a probability of 30\% - 50\%. We trained the model with the encoder pre-loaded with Imagenet weights, using Adam optimizer with a learning rate of 0.00001 and a weight decay of 0.00005. Learning rate scheduling was done using Reduction on Plateau by a factor of 0.25. All models were trained for 60 epochs and with Early-Stopping patience of 20 epochs. The model was implemented using PyTorch. For the EfficinetNet backbone
we used the implementation from Timm models \cite{rw2019timm}.

\begin{figure}
        \centering
        \begin{subfigure}[b]{\linewidth}
            \centering
            \includegraphics[width=\textwidth]{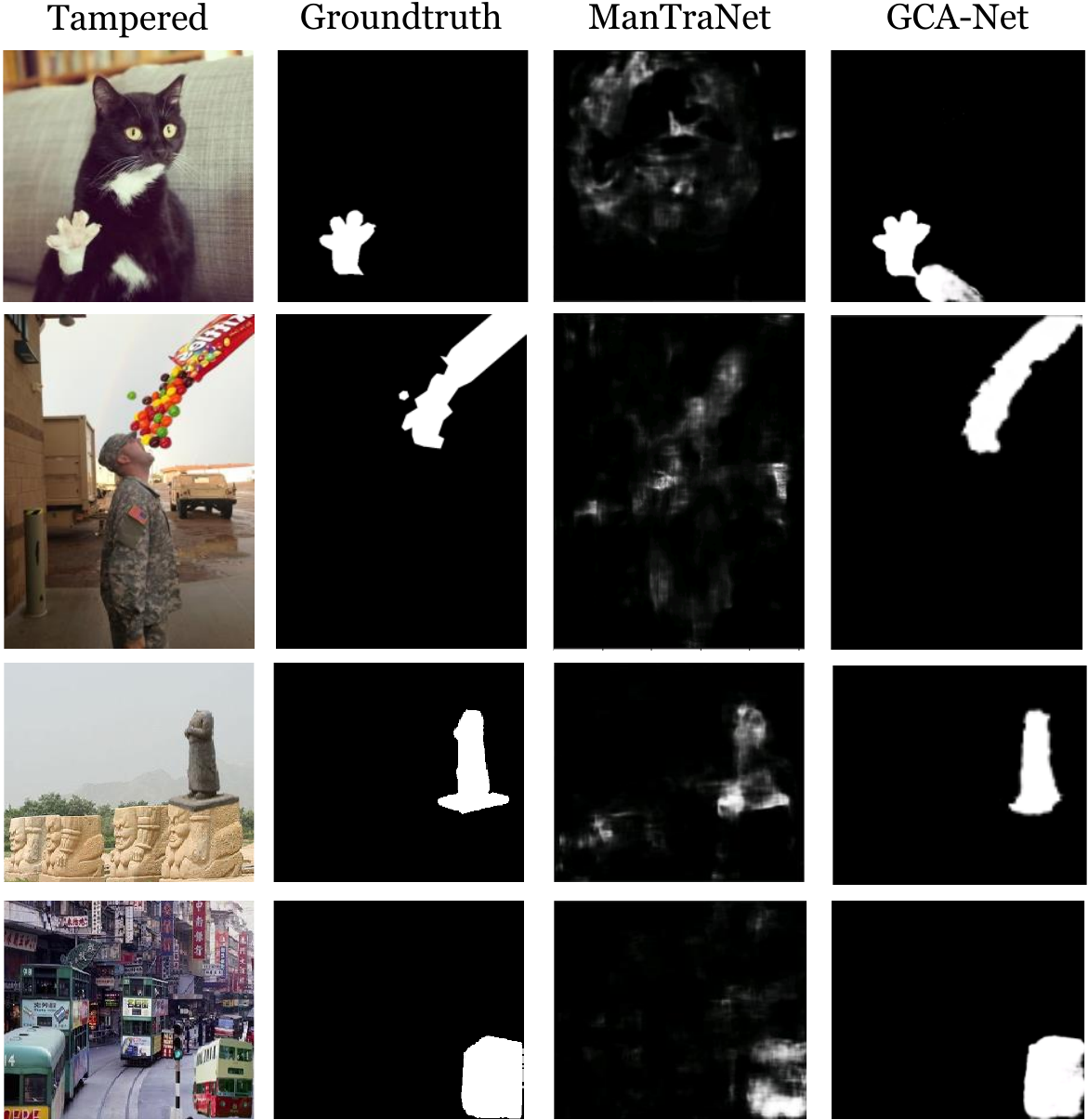}
        \end{subfigure}
        \vskip\baselineskip
        \begin{subfigure}[b]{\linewidth}   
            \centering 
            \includegraphics[width=\textwidth]{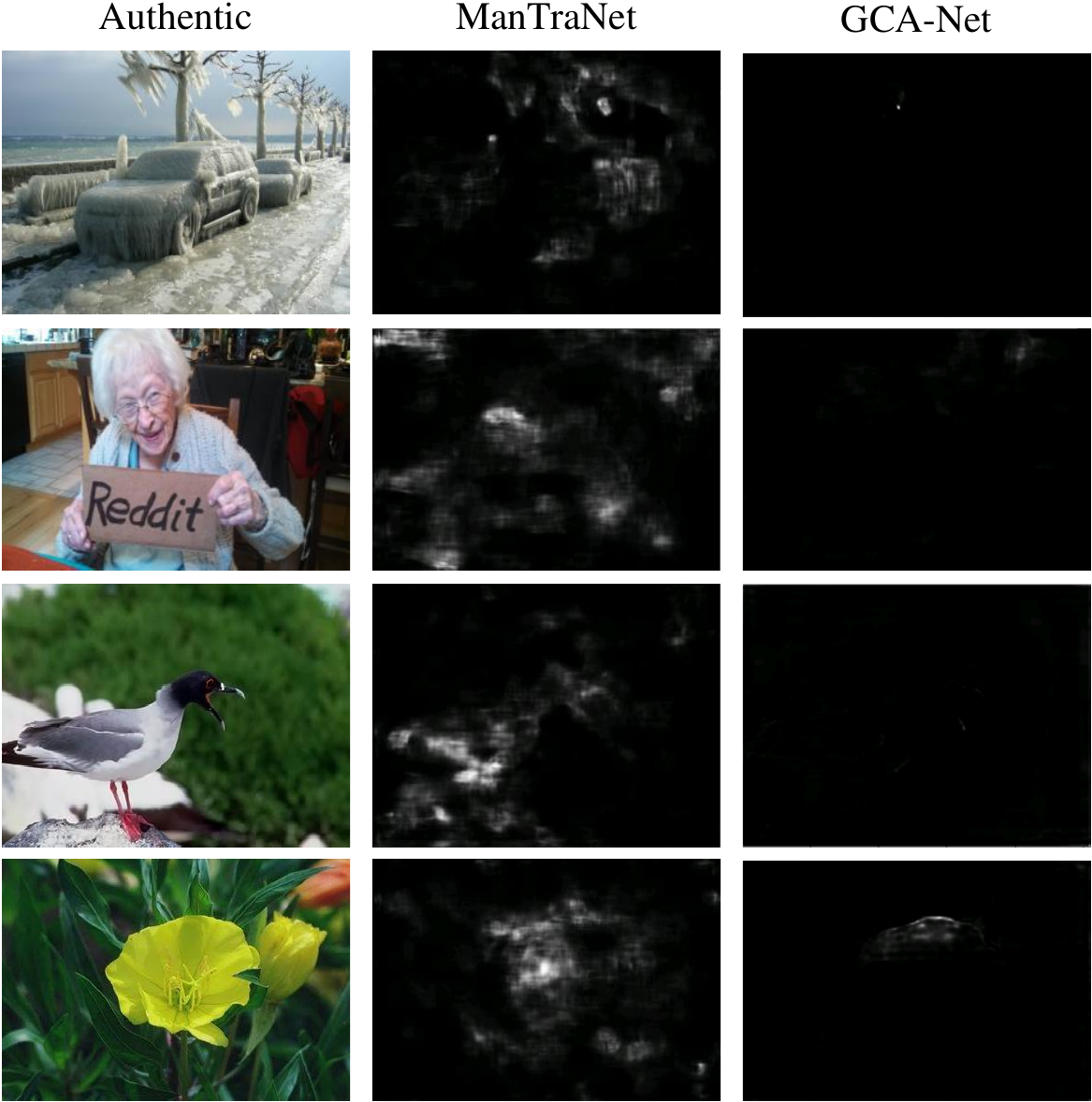}
        \end{subfigure}
        \caption{ Qualitative comparison of GCA-Net and ManTraNet for various tampered and authentic images.} 
        \label{fig:qual}
\end{figure}

{\small
\bibliographystyle{ieee_fullname}
\bibliography{egbib}

\begin{thebibliography}{10}\itemsep=-1pt

\bibitem{lstm}
Jawadul~H Bappy, Cody Simons, Lakshmanan Nataraj, BS Manjunath, and Amit~K
  Roy-Chowdhury.
\newblock Hybrid lstm and encoder-decoder architecture for detection of image
  forgeries.
\newblock {\em IEEE Transactions on Image Processing}, 2019.

\bibitem{detect_survey}
Z.~J. {Barad} and M.~M. {Goswami}.
\newblock Image forgery detection using deep learning: A survey.
\newblock In {\em 2020 6th International Conference on Advanced Computing and
  Communication Systems (ICACCS)}, pages 571--576, 2020.

\bibitem{constrained}
B. {Bayar} and M.~C. {Stamm}.
\newblock Constrained convolutional neural networks: A new approach towards
  general purpose image manipulation detection.
\newblock {\em IEEE Transactions on Information Forensics and Security}, 2018.

\bibitem{maskrcnn}
T.~Aaron Gulliver \& Saif~alZahir Belal~Ahmed.
\newblock Image splicing detection using mask-rcnn.
\newblock {\em Signal, Image and Video Processing}, 2020.

\bibitem{rru}
X. {Bi}, Y. {Wei}, B. {Xiao}, and W. {Li}.
\newblock Rru-net: The ringed residual u-net for image splicing forgery
  detection.
\newblock In {\em 2019 IEEE/CVF Conference on Computer Vision and Pattern
  Recognition Workshops (CVPRW)}, pages 30--39, 2019.

\bibitem{imanip_survey}
Gajanan~K. Birajdar and Vijay~H. Mankar.
\newblock Digital image forgery detection using passive techniques: A survey.
\newblock {\em Digital Investigation}, 10(3):226 -- 245, 2013.

\bibitem{camera1}
L. {Bondi}, S. {Lameri}, D. {Güera}, P. {Bestagini}, E.~J. {Delp}, and S.
  {Tubaro}.
\newblock Tampering detection and localization through clustering of
  camera-based cnn features.
\newblock In {\em 2017 IEEE Conference on Computer Vision and Pattern
  Recognition Workshops (CVPRW)}, pages 1855--1864, 2017.

\bibitem{gcnet}
Yue Cao, Jiarui Xu, Stephen Lin, Fangyun Wei, and Han Hu.
\newblock Gcnet: Non-local networks meet squeeze-excitation networks and
  beyond, 2019.

\bibitem{chen2017deeplab}
Liang-Chieh Chen, George Papandreou, Iasonas Kokkinos, Kevin Murphy, and
  Alan~L. Yuille.
\newblock Deeplab: Semantic image segmentation with deep convolutional nets,
  atrous convolution, and fully connected crfs, 2017.

\bibitem{multi-scale}
Xinru Chen, Chengbo Dong, Jiaqi Ji, Juan Cao, and Xirong Li.
\newblock Image manipulation detection by multi-view multi-scale supervision,
  2021.

\bibitem{prnu2}
Giovanni Chierchia, Giovanni Poggi, Carlo Sansone, and Luisa Verdoliva.
\newblock Prnu-based forgery detection with regularity constraints and global
  optimization.
\newblock In {\em 2013 IEEE 15th International Workshop on Multimedia Signal
  Processing (MMSP)}, 2013.

\bibitem{prnu1}
Giovanni Chierchia, Giovanni Poggi, Carlo Sansone, and Luisa Verdoliva.
\newblock A bayesian-mrf approach for prnu-based image forgery detection.
\newblock {\em IEEE Transactions on Information Forensics and Security}, 9,
  2014.

\bibitem{chollet2017xception}
François Chollet.
\newblock Xception: Deep learning with depthwise separable convolutions, 2017.

\bibitem{noiseprint}
Davide Cozzolino and Luisa Verdoliva.
\newblock Noiseprint: a cnn-based camera model fingerprint, 2018.

\bibitem{ontheface}
Hao Dang, Feng Liu, Joel Stehouwer, Xiaoming Liu, and Anil~K. Jain.
\newblock On the detection of digital face manipulation.
\newblock In {\em Proceedings of the IEEE/CVF Conference on Computer Vision and
  Pattern Recognition (CVPR)}, June 2020.

\bibitem{casia}
Jing Dong, Wei Wang, and Tieniu Tan.
\newblock Casia image tampering detection evaluation database.
\newblock pages 422--426, 07 2013.

\bibitem{cfai}
P. {Ferrara}, T. {Bianchi}, A. {De Rosa}, and A. {Piva}.
\newblock Image forgery localization via fine-grained analysis of cfa
  artifacts.
\newblock {\em IEEE Transactions on Information Forensics and Security},
  7(5):1566--1577, 2012.

\bibitem{ghosh2019spliceradar}
Aurobrata Ghosh, Zheng Zhong, Terrance~E Boult, and Maneesh Singh.
\newblock Spliceradar: A learned method for blind image forensics, 2019.

\bibitem{dresden}
Thomas Gloe and Rainer B\"{o}hme.
\newblock The 'dresden image database' for benchmarking digital image
  forensics.
\newblock Association for Computing Machinery, 2010.

\bibitem{nist}
H. {Guan}, M. {Kozak}, E. {Robertson}, Y. {Lee}, A.~N. {Yates}, A. {Delgado},
  D. {Zhou}, T. {Kheyrkhah}, J. {Smith}, and J. {Fiscus}.
\newblock Mfc datasets: Large-scale benchmark datasets for media forensic
  challenge evaluation.
\newblock In {\em 2019 IEEE Winter Applications of Computer Vision Workshops
  (WACVW)}, pages 63--72, 2019.

\bibitem{squeezeandexcitation}
Jie Hu, Li Shen, Samuel Albanie, Gang Sun, and Enhua Wu.
\newblock Squeeze-and-excitation networks, 2019.

\bibitem{span}
Xuefeng Hu, Zhihan Zhang, Zhenye Jiang, Syomantak Chaudhuri, Zhenheng Yang, and
  Ram Nevatia.
\newblock Span: Spatial pyramid attention network for image manipulation
  localization.
\newblock In Andrea Vedaldi, Horst Bischof, Thomas Brox, and Jan-Michael Frahm,
  editors, {\em Computer Vision -- ECCV 2020}. Springer International
  Publishing, 2020.

\bibitem{densenet}
Gao Huang, Zhuang Liu, Laurens van~der Maaten, and Kilian~Q. Weinberger.
\newblock Densely connected convolutional networks, 2018.

\bibitem{faceswap}
Iryna Korshunova, Wenzhe Shi, Joni Dambre, and Lucas Theis.
\newblock Fast face-swap using convolutional neural networks, 2016.

\bibitem{ELA2}
N. Krawetz.
\newblock A picture ’ s worth . . . digital image analysis and forensics
  version 2.
\newblock 2007.

\bibitem{inp1}
Jingyuan Li, Fengxiang He, Lefei Zhang, Bo Du, and Dacheng Tao.
\newblock Progressive reconstruction of visual structure for image inpainting.
\newblock In {\em Proceedings of the IEEE/CVF International Conference on
  Computer Vision (ICCV)}, October 2019.

\bibitem{focal}
Tsung-Yi Lin, Priya Goyal, Ross Girshick, Kaiming He, and Piotr Dollár.
\newblock Focal loss for dense object detection, 2018.

\bibitem{improved_jpeg}
Huajian Liu, Martin Steinebach, and Kathrin Schölei.
\newblock Improved manipulation detection with convolutional neural network for
  jpeg images.
\newblock pages 1--6, 08 2019.

\bibitem{pscc}
Xiaohong Liu, Yaojie Liu, Jun Chen, and Xiaoming Liu.
\newblock Pscc-net: Progressive spatio-channel correlation network for image
  manipulation detection and localization, 2021.

\bibitem{wavelet}
Babak Mahdian and Stanislav Saic.
\newblock Using noise inconsistencies for blind image forensics.
\newblock {\em Image Vision Comput.}, 27(10), 2009.

\bibitem{defacto}
Gaël MAHFOUDI, Badr TAJINI, Florent RETRAINT, Frédéric MORAIN-NICOLIER,
  Jean~Luc DUGELAY, and Marc PIC.
\newblock Defacto: Image and face manipulation dataset.
\newblock In {\em 2019 27th European Signal Processing Conference (EUSIPCO)},
  2019.

\bibitem{skip-connection}
Ghazal Mazaheri, Niluthpol Chowdhury~Mithun, Jawadul~H. Bappy, and Amit~K.
  Roy-Chowdhury.
\newblock A skip connection architecture for localization of image
  manipulations.
\newblock In {\em Proceedings of the IEEE/CVF Conference on Computer Vision and
  Pattern Recognition (CVPR) Workshops}, June 2019.

\bibitem{siamese}
Aniruddha Mazumdar, Jaya Singh, Yosha~Singh Tomar, and Prabin~Kumar Bora.
\newblock Universal image manipulation detection using deep siamese
  convolutional neural network, 2018.

\bibitem{local1}
G. {Muzaffer} and G. {Ulutas}.
\newblock A new deep learning-based method to detection of copy-move forgery in
  digital images.
\newblock In {\em 2019 Scientific Meeting on Electrical-Electronics Biomedical
  Engineering and Computer Science (EBBT)}, 2019.

\bibitem{imd2020}
A. {Novozámský}, B. {Mahdian}, and S. {Saic}.
\newblock Imd2020: A large-scale annotated dataset tailored for detecting
  manipulated images.
\newblock In {\em 2020 IEEE Winter Applications of Computer Vision Workshops
  (WACVW)}, pages 71--80, 2020.

\bibitem{attention-unet}
Ozan Oktay, Jo Schlemper, Loic~Le Folgoc, Matthew Lee, Mattias Heinrich,
  Kazunari Misawa, Kensaku Mori, Steven McDonagh, Nils~Y Hammerla, Bernhard
  Kainz, Ben Glocker, and Daniel Rueckert.
\newblock Attention u-net: Learning where to look for the pancreas, 2018.

\bibitem{dense_camera}
Abdul~Muntakim Rafi, Uday Kamal, Rakibul Hoque, Abid Abrar, Sowmitra Das,
  Robert Laganière, and Md.~Kamrul Hasan.
\newblock Application of densenet in camera model identification and
  post-processing detection, 2019.

\bibitem{detect1}
Yuan Rao and Jiangqun Ni.
\newblock A deep learning approach to detection of splicing and copy-move
  forgeries.
\newblock 2016.

\bibitem{scse}
Abhijit~Guha Roy, Nassir Navab, and Christian Wachinger.
\newblock Recalibrating fully convolutional networks with spatial and channel
  'squeeze \& excitation' blocks, 2018.

\bibitem{faceforensics}
Andreas Rössler, Davide Cozzolino, Luisa Verdoliva, Christian Riess, Justus
  Thies, and Matthias Nießner.
\newblock Faceforensics++: Learning to detect manipulated facial images, 2019.

\bibitem{reducedfocal}
Nikolay Sergievskiy and Alexander Ponamarev.
\newblock Reduced focal loss: 1st place solution to xview object detection in
  satellite imagery, 2019.

\bibitem{attributechanging}
Yujun Shen, Jinjin Gu, Xiaoou Tang, and Bolei Zhou.
\newblock Interpreting the latent space of gans for semantic face editing,
  2020.

\bibitem{dual_domain}
Z. {Shi}, X. {Shen}, H. {Kang}, and Y. {Lv}.
\newblock Image manipulation detection and localization based on the
  dual-domain convolutional neural networks.
\newblock {\em IEEE Access}, 2018.

\bibitem{vgg}
Karen Simonyan and Andrew Zisserman.
\newblock Very deep convolutional networks for large-scale image recognition,
  2015.

\bibitem{noise1}
Xunyu Pan \& Xing~Zhang Siwei~Lyu.
\newblock Exposing region splicing forgeries with blind local noise estimation.
\newblock {\em International Journal of Computer Vision}, 2014.

\bibitem{dice}
Carole~H. Sudre, Wenqi Li, Tom Vercauteren, Sebastien Ourselin, and M.
  Jorge~Cardoso.
\newblock Generalised dice overlap as a deep learning loss function for highly
  unbalanced segmentations.
\newblock {\em Lecture Notes in Computer Science}, 2017.

\bibitem{tan2020efficientnet}
Mingxing Tan and Quoc~V. Le.
\newblock Efficientnet: Rethinking model scaling for convolutional neural
  networks, 2020.

\bibitem{detetct4}
Kunj Bihari Meena \&~Vipin Tyagi.
\newblock A hybrid copy-move image forgery detection technique based on
  fourier-mellin and scale invariant feature transforms.
\newblock {\em Multimedia Tools and Applications}, 2020.

\bibitem{attenionisall}
Ashish Vaswani, Noam Shazeer, Niki Parmar, Jakob Uszkoreit, Llion Jones,
  Aidan~N. Gomez, Lukasz Kaiser, and Illia Polosukhin.
\newblock Attention is all you need.
\newblock 2017.

\bibitem{jpeg_noise}
Vinay Verma, Deepak Singh, and Nitin Khanna.
\newblock Block-level double jpeg compression detection for image forgery
  localization, 2020.

\bibitem{non-local}
Xiaolong Wang, Ross Girshick, Abhinav Gupta, and Kaiming He.
\newblock Non-local neural networks.
\newblock In {\em Proceedings of the IEEE Conference on Computer Vision and
  Pattern Recognition (CVPR)}, June 2018.

\bibitem{ela}
N.~B.~A. {Warif}, M.~Y.~I. {Idris}, A.~W.~A. {Wahab}, and R. {Salleh}.
\newblock An evaluation of error level analysis in image forensics.
\newblock In {\em 2015 5th IEEE International Conference on System Engineering
  and Technology (ICSET)}, pages 23--28, 2015.

\bibitem{rcnn2}
Liqing Wei, Wu, Dong, Zhang, and Sun.
\newblock Developing an image manipulation detection algorithm based on edge
  detection and faster r-cnn.
\newblock {\em Symmetry}, 11:1223, 10 2019.

\bibitem{coverage}
B. {Wen}, Y. {Zhu}, R. {Subramanian}, T. {Ng}, X. {Shen}, and S. {Winkler}.
\newblock Coverage — a novel database for copy-move forgery detection.
\newblock In {\em 2016 IEEE International Conference on Image Processing
  (ICIP)}, pages 161--165, 2016.

\bibitem{rw2019timm}
Ross Wightman.
\newblock Pytorch image models.
\newblock \url{https://github.com/rwightman/pytorch-image-models}, 2019.

\bibitem{deepmatching}
Yue Wu, Wael AbdAlmageed, and Prem Natarajan.
\newblock Deep matching and validation network -- an end-to-end solution to
  constrained image splicing localization and detection, 2017.

\bibitem{busternet}
Yue Wu, Wael AbdAlmageed, and Prem Natarajan.
\newblock Busternet: Detecting image copy-move forgery with source/target
  localization.
\newblock In {\em European Conference on Computer Vision (ECCV)}. Springer,
  2018.

\bibitem{mantranet}
Y. {Wu}, W. {AbdAlmageed}, and P. {Natarajan}.
\newblock Mantra-net: Manipulation tracing network for detection and
  localization of image forgeries with anomalous features.
\newblock In {\em 2019 IEEE/CVF Conference on Computer Vision and Pattern
  Recognition (CVPR)}, pages 9535--9544, 2019.

\bibitem{resnext}
Saining Xie, Ross Girshick, Piotr Dollár, Zhuowen Tu, and Kaiming He.
\newblock Aggregated residual transformations for deep neural networks, 2017.

\bibitem{nlp-gate}
Lanqing Xue, Xiaopeng Li, and Nevin~L. Zhang.
\newblock Not all attention is needed: Gated attention network for sequence
  data, 2019.

\bibitem{yang2020constrained}
Chao Yang, Huizhou Li, Fangting Lin, Bin Jiang, and Hao Zhao.
\newblock Constrained r-cnn: A general image manipulation detection model,
  2020.

\bibitem{graph-gate}
Jiani Zhang, Xingjian Shi, Junyuan Xie, Hao Ma, Irwin King, and Dit-Yan Yeung.
\newblock Gaan: Gated attention networks for learning on large and
  spatiotemporal graphs, 2018.

\bibitem{zhou2019generate}
Peng Zhou, Bor-Chun Chen, Xintong Han, Mahyar Najibi, Abhinav Shrivastava,
  Ser~Nam Lim, and Larry~S. Davis.
\newblock Generate, segment and refine: Towards generic manipulation
  segmentation, 2019.

\bibitem{rich_features}
Peng Zhou, Xintong Han, Vlad~I. Morariu, and Larry~S. Davis.
\newblock Learning rich features for image manipulation detection, 2018.

\bibitem{unet++}
Zongwei Zhou, Md~Mahfuzur~Rahman Siddiquee, Nima Tajbakhsh, and Jianming Liang.
\newblock Unet++: A nested u-net architecture for medical image segmentation,
  2018.

\end{thebibliography}
}


\end{document}